# Land use mapping in the Three Gorges Reservoir Area based on semantic segmentation deep learning method

**Xin Zhang, Bingfang Wu\*, Liang Zhu, Fuyou Tian, Miao Zhang and Yuanzeng**

Key Laboratory of Digital Earth Science, Institute of Remote Sensing and Digital Earth, Chinese Academy of Sciences, Olympic Village Science Park, W. Beichen Road, Beijing 100101, China;

\*Correspondence: wubf@radi.ac.cn; Tel.: +86[-010-6485-8721]

## Abstract:

The Three Gorges Dam, a massive cross-century project spans the Yangtze River by the town of Sandouping, located in Yichang, Hubei province, China, was built to provide great power, improve the River shipping, control floods in the upper reaches of the Yangtze River, and increase the dry season flow in the middle and lower reaches of the Yangtze River. Benefits are enormous and comprehensive. However, the social and environmental impacts are also immense and far-reaching to its surrounding areas. Mapping land use /land cover changed (LUCC) is critical for tracking the impacts. Remote sensing has been proved to be an effective way to map and monitor land use change in real time and in large areas such as the Three Gorges Reservoir Area(TGRA) by using pixel based or oriented based classifier in different resolution. In this paper, we first test the state of the art semantic segmentation deep learning classifiers for LUCC mapping with 7 categories in the TGRA area with rapideye 5m resolution data. The topographic information was also added for better accuracy in mountain area. By compared with the pixel-based classifier, the semantic segmentation deep learning method has better accuracy and robustness at 5m resolution level.

# Introduction:

The Three Gorges Project (TGP) involved building the largest dam on the third longest river in the world. The TGP has created a reservoir area of 1,080 km$^2$ by damming the Yangtze River and has greatly changed the landscape pattern in the Three Gorges Reservoir Area (TGRA)(Zhang et al. 2012). Large dams provide many benefits to the society but simultaneously cause adverse and often irreversible impacts on the environment(Zhang et al. 2009; Tullos 2009).

Over the past 60 years, several studies have described the impacts of Egypt's Aswan High Dam (once the world's largest dam) on its surrounding environment, aspects include ecology, water and soil quality and human health(Zeid 1989; Moussa et al. 2013). The influence of the world's second largest hydroelectric dam, the Itaipu hydroelectric project, located along the border river between Brazil and Paraguay, has also been widely studied(Stivari et al. 2005; Murphy 1976). As the world's largest hydroelectric project today, the TGP plays a crucial role in flood prevention and control, power generation (with a total capacity of 18,200 MW) and shipping(Yang et al. 2014). However, the dam will continuously cause some adverse changes with respect to the natural environment, land resources, and even socio-economic development in the TGRA and related areas.

To characterize the environmental effects and minimize the adverse influences associated from the TGP, the Chinese government has gradually formulated and implemented a series of relevant polices(Wu et al. 2004; Tullos 2009; Zhang and Lou 2011). In 1996, the Three Gorges Project Construction Committee of the State Council set up the "Three Gorges Project Ecological and Environmental Monitoring System," which is responsible for monitoring the ecological and environmental problems related to the Three Gorges Project. Land cover properties and structure contribute substantially to regional environmental changes(Lindquist et al. 2008). Therefore, land use/land cover changes (LUCC) have been recognized as the most important indicator to study ecological and environmental changes(Pabi 2007; Meyer and Turner 1995; West 1995).

Remote sensing has proven to be the most effective method for monitoring LUCC at large scales using temporal and spectral information(Gong et al. 2013; Mora et al. 2014).For many years, traditional classifications of land cover based on remote sensing imagery have resorted to two main methods: pixel-based and object-based(Zhang et al. 2014). Pixel-based classification methods use algorithms such as maximum likelihood maximum likelihood(Strahler 1980), minimum distance and logistic regression(Bondell 2005), artificial neural networks(Van Coillie et al. 2011), support vector machines(Foody and Mathur 2004), random forest(Pal 2005), and K nearest neighbor(Falkowski et al. 2010). The pixel-based method is often used with hyperspectral or time-series data to achieve high accuracy because multiple features can be provided for each cell(Jimenezrodriguez 1999; Chen et al. 2014). However, increases in within-class variance and decreases in between-class variance often lead to the inadequacies of traditional pixel-based classification approaches(Huang and Zhang 2012). In addition, with the increase in the spatial resolution of sensors to 5-30 m resolution or higher, the pixel-based classification results tend to possess "salt and pepper" noise, that is, in some categories there are other categories of noise due to the effect of spectral confusion among different land cover classes(Chan et al. 2005). To overcome the problem of salt and pepper noise, object-oriented classification began to be applied to high-resolution images. In contrast to the pixel-based classification method, object-based classification uses both spectral and spatial features by dividing traditional classification into two steps: segmentation and classification(Ding 2005). Image segmentation is a principal function that splits an image into separate regions or objects depending on specified parameters. A group of pixels with similar spectral and spatial properties is considered to be an object in the object-based classification prototype. The segmentation techniques utilize spatial concepts that involve geometric features, spatial relations, and scale topology relations of upscale and downscale inheritances(Benz et al. 2004; Burnett and Blaschke 2003). The classical segmentation algorithms mainly include the recursive hierarchical segmentation (RHSeg) segmentation algorithm(Tilton 1998; Adams and Bischof 1994), multiresolution segmentation(Baatz and Schäpe 2000) and watershed segmentation(Roerdink et al. 2000). The properties of each object are assigned according to spectral features. With the use of object-based classification, similar pixels are aggregated into an object via segmentation, thereby avoiding salt and pepper noise.

Over the past few years, deep learning has become a technology of considerable interest for computer vision tasks that rely on the rapid development of Graphic Processing Units (GPU). Convolution neural networks (CNNs) are very powerful tools that yield hierarchies of features and are now the state of the art for object recognition and classification(Audebert et al. 2016). Recently, deep networks based on CNNs have been applied to semantic pixel-wise segmentation, which consists in assigning a semantic label (i.e., a class) to each coherent region of an image. By using pixel-wise dense prediction models, semantic segmentation models can classify each pixel in an image.

Fully convolutional networks (FCNs) were first used for the semantic segmentation of natural images(Long et al. 2015), multi-modal medical image analyses and multispectral satellite image segmentation. SegNet(Badrinarayanan et al. 2015), U-Net(Ronneberger et al. 2015), FC-Densenet(Jégou et al. 2016), E-Net(Paszke et al. 2016), Link-Net(Chaurasia and Culurciello 2017), RefineNet(Lin et al. 2016) and PSPNet(Zhao et al. 2016) were designed to perform semantic segmentation.

In this paper, We assessed the feasibility of the SegNet model for remote sensing data classification by attempting to set different overlaps on training sample and bands combination inputs.

# Study area

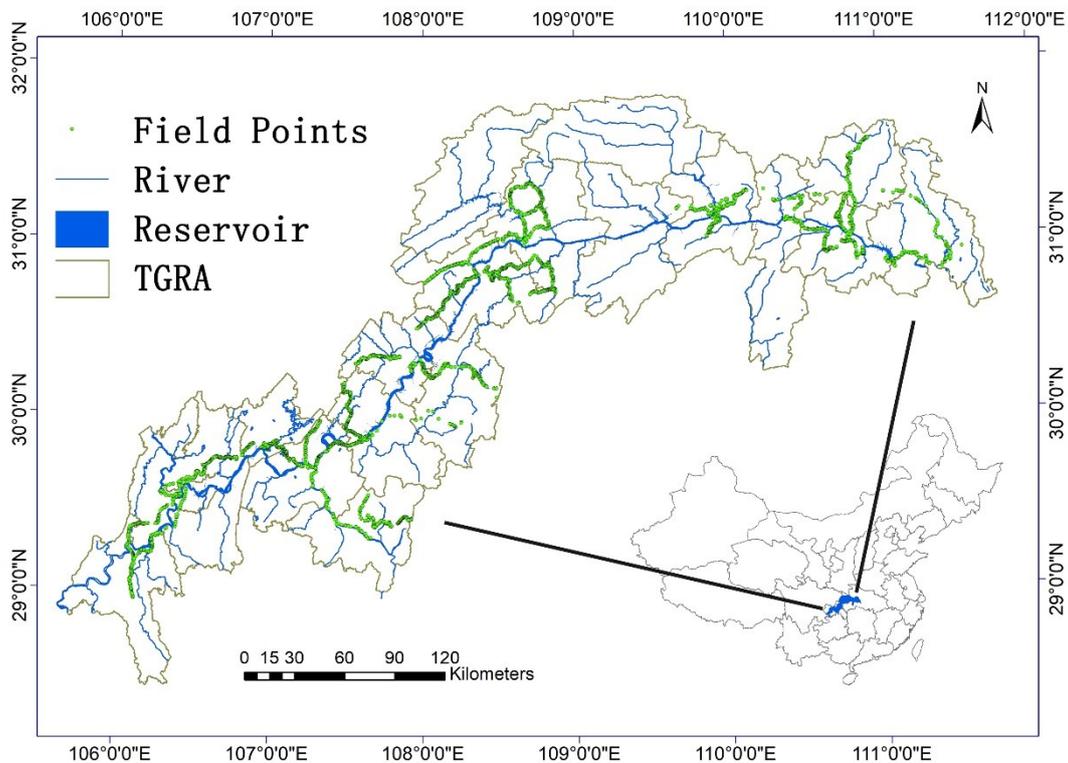

Figure 1 Location map of the Three Gorges Reservoir Area（TGRA）

The TGRA study area is in the transition zone from the Qinghai-Tibet Plateau to the lower reaches of the Yangtze River between 106°16′E–111°28′E and 28°56′N–31°44′N (Figure 1). The TGRA covers the lower section of the upper Yangtze and has an area of 58,000 km$^2$ and a population of almost 20 million. Approximately 74% of the region is mountainous, 4.3% are plains, and 21.7% is hilly, with elevations between 73 and 2,977 m. The region consists of 20 counties or districts, sixteen of which are in Chongqing Province and four in Hubei Province. The study area is ecologically vulnerable due to height differences, which produce frequent landslides. Due to the monsoon climate, there is obvious seasonality in the TGRA. Annual precipitation ranges from 1,000 to 1,200 mm, with 60% of the rainfall occurring between June and September. Due to impacts arising from the dam, approximately 1.25 million people were displaced over a 16-year period ending in 2008. The inundation of such an enormous land area represents a major land cover change.

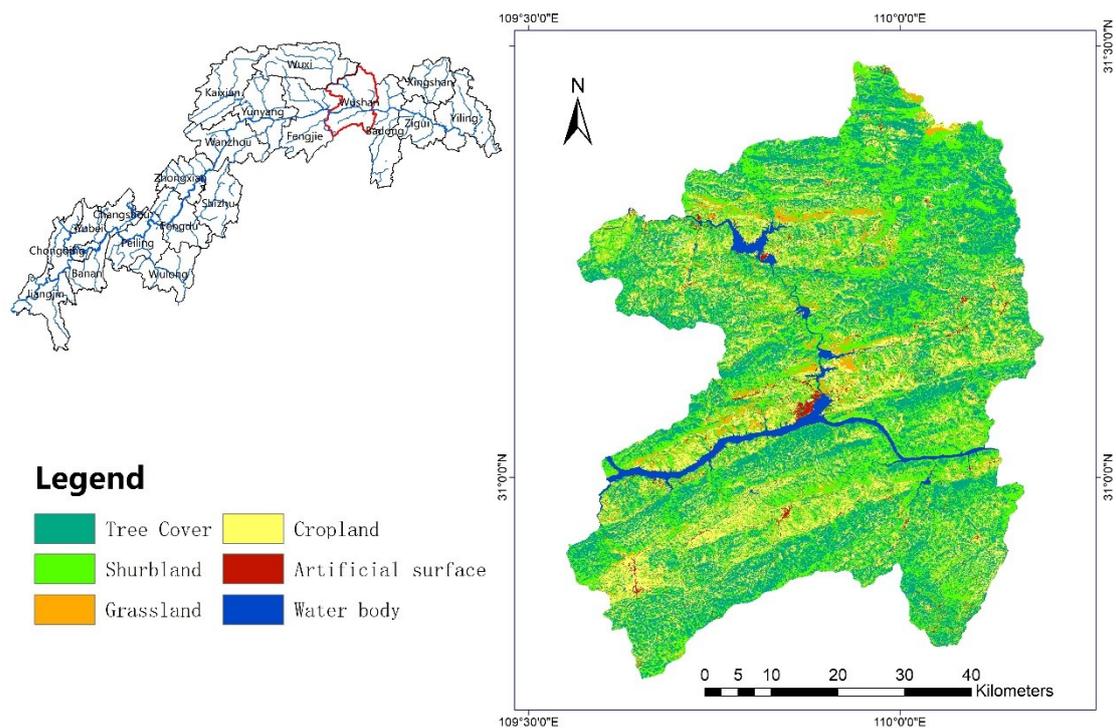

Figure 2 Wushan county with its LULC map

Wushan County located at the western entrance to the Wu Gorge in the TGRA is always called as the 'core heart' of TGRA. The location is between 109°33′E–110°11′E and 30°45′N–23°28′N. Wushan occupies approximately 2,958 km² and has a population of approximately 600,000. In this study, training a model in the entire TGRA required approximately one month for 400, 000 iterations. We choose Wushan which is a smaller typical area to quantify the model performance of different training inputs.

# Data and methods

## Data

### Remote sensing data

Spectral information has always been the most important component to identify features in remote sensing imagery(Richards 2006). The RapidEye constellation was launched into orbit on August 29, 2008. The five satellites, designated herein as RE1, RE2, RE3, RE4, and RE5, are phased in a sun-synchronous orbit plane with an inclination of 97.8°and a local time at ascending node (LTAN) of 23:30. The RapidEye constellation sensors provided 5-meter resolution imagery in five-bands: blue (440–510 nm), green (520–590 nm), red (630–685 nm), red edge (690–730 nm) and near infrared (760–850 nm). These sensors provided abundant spatial and spectral information, which have been widely used for agriculture, forestry, environmental monitoring and other applications. The series-unique red-edge band, which is sensitive to changes in chlorophyll content, can assist in monitoring vegetation health, improving species separation and measuring protein and nitrogen content in biomass. For this paper, 192 of 2012 and 198 of 2016 tiles with the minimum cloud cover in summer and autumn were selected in this paper.

## Auxiliary data

In addition, topographic information such as elevation, slope and aspect can provide meaningful information for land cover recognition. Benediktsson demonstrated that the elevation source is responsible for up to 80% of the weight on the land cover classification(Benediktsson et al. 1990). A DEM and 5-m slope data, which provided crucial information for land-use classification, were also used as auxiliary data. Slopes were calculated from elevation using the ARCGIS Spatial Analyst toolbox(Burrough et al. 1986).

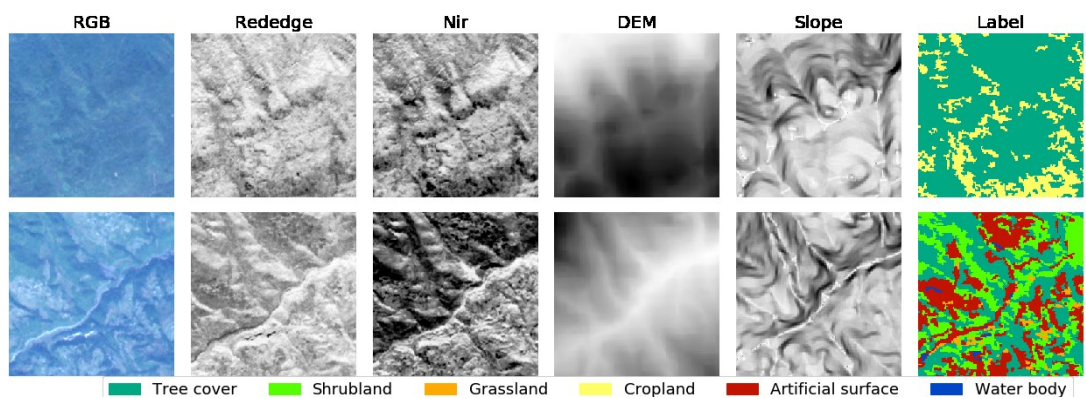

## Training sample and ground truth

The training labels were built by 2012 land cover. The data were obtained from RapidEye imagery using object-oriented classification methods, including multiresolution segmentation(Benz et al. 2004) and the classification and regression trees (CART)(Fonarow et al. 2005) classifier developed by eCognition(Ding 2005). The results of the classification were validated via extensive ground survey and massive manual correction. The average accuracy of the 7 classes reached 86%.

The ground survey points were collected using GVG (a volunteered geographic information (VGI) smartphone app), which can be downloaded from the iTunes app store. In this work, 4,174 points obtained in 2016 were used for evaluating the precision of the predictions in the TGRA.

# Methods

## 1) Overview of the Methodology

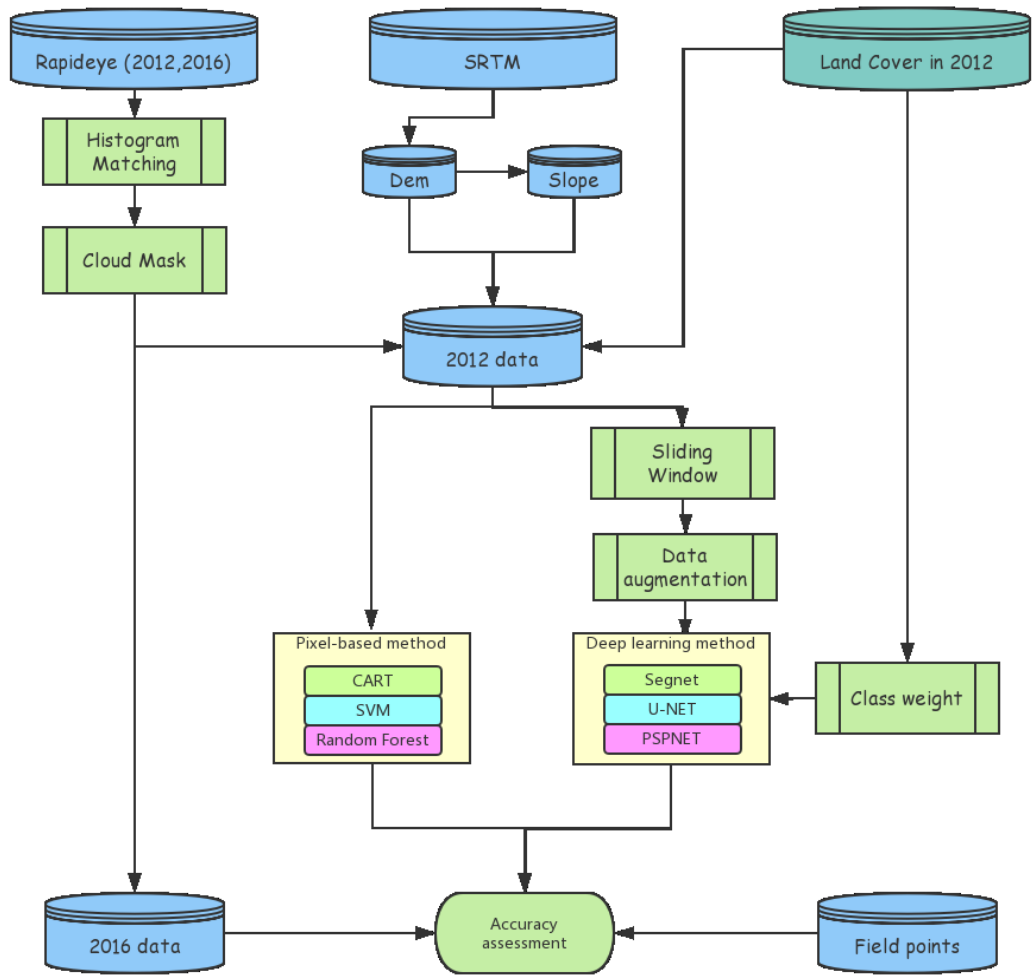

Figure 3 flow chart

A comprehensive overview of the methodology is shown in FigureX. The land cover classes in this work are mainly based on the land cover classification system of the Intergovernmental Panel on Climate Change (IPCC) including Forest, Shrub, Grassland, Farmland, Building-up and Wetland. The rapideye data after histogram matching and cloud mask with topographic information (DEM and slope) and their landcover in 2012 were used for training the classifier. Three most popular pixel-based classifiers including Classification and Regression Tree (CART), Support Vector Machine (SVM) and Random Forest (RF) and three significant semantic segmentation deep learning classifier (Segnet, U-NET and PSPNET) were tested in this work. For pixel-based classifier, XXXX random points for each type by using stratified sampling based on their geolocation were used for training, the detailed parameters are shown on table XX. For deep learning classifier, 60% of the samples after samples were used for training the model, the 20% of the samples was used to validate the training model. The left 20% of the samples to evaluate all the classifiers in this work. Besides, we applied the best pixel-based classifier and deep learning classifier in the data of 2016. XXX field points were used to evaluate the generalization and adaptability on new data.

## Data pre-processing

## Histogram matching

Images with minimal cloud cover and approximate acquisition times were selected for training and prediction in 2012 and 2016. Due to different atmospheric and lighting conditions during the different acquisition times, obvious brightness differences for each image remained. To align the apparent distributions of the brightness values in the images as closely as possible, histogram matching was used to adjust and mosaic all the images(Richards 2006). The original image and the images after histogram matching are shown in Figure 4.

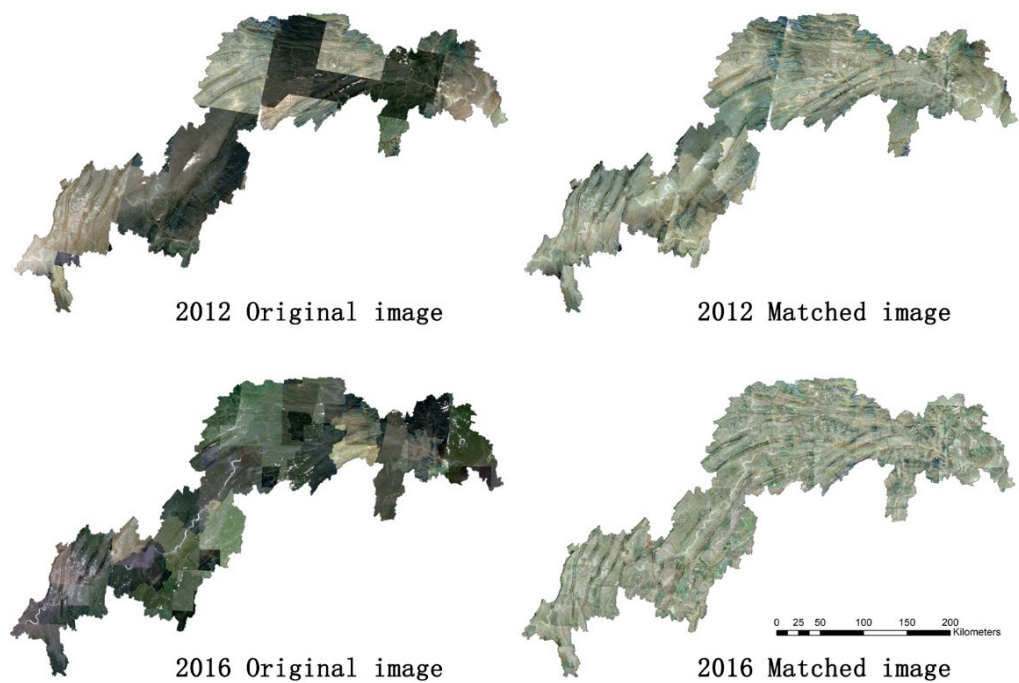

Figure 4 RapidEye data coverage in the TGRA. The original image is obtained by mosaicking all the original tiles directly. The matched image is obtained by mosaicking the images after histogram matching

## Cloud masks

Although images with minimal cloud cover were selected for training, clouds still affected the training results. We used RapidEye cloud products to label all the cloud pixels as Nodata during the training.

## Sliding window

Remote sensing images are too large to pass through a CNN, and most CNNs are tailored for a resolution of 256 × 256 or 512 × 512 pixels. Given current GPU memory limitations, we split our training data into smaller patches using a simple sliding window (Figure 5). Meanwhile, to adjust the overlap rate can also increase the number of sample which can reduce the overfitting. In this study, the original image are divided into 256 × 256 patches and 512 × 512 patches respectively. The overlap rate was set to half of the patches size which can increase 0.5 times of samples.

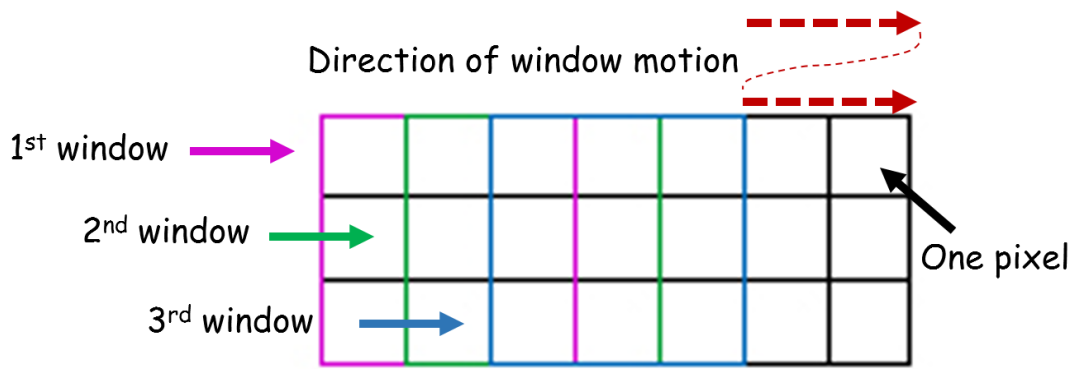

Figure 5 Sliding window method in this study, a 256*256 or 512 *512 pixels moving window slide through the image with a setting distance (half of the sample size) each time to get a sample.

## Data augmentation

The easiest and most common method to reduce overfitting on image data is to artificially enlarge the dataset using label-preserving transformations(Krizhevsky et al. 2012). In this study, four forms of data augmentation were used for each sample, Rotate -90⁰ and 90⁰, flip and mirror (Figure 6).

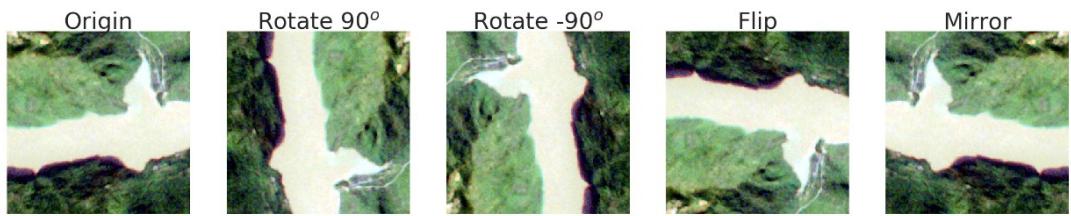

Figure 6

# Class weight

In the natural environment, the area of different types is different and vary greatly which made the classes unbalanced for training the deep learning model. a class weight was used for weighting the loss function (during training only). This can be useful to tell the model to "pay more attention" to samples from an under-represented class(Gary and Zeng 2001). The class weight for each type was given by the reciprocal of the proportion of the total area occupied by this type.

## Pixel based classification methods

Non-parametric supervised classifiers were widely used in remote sensing imagery classification task, such as the Classification and Regression Tree (CART)(Hua et al. 2017; Chasmer et al. 2014; Friedl and Brodley 1997), Support Vector Machine (SVM)(Munoz-Mari et al. 2010; Melgani and Bruzzone 2004; Mountrakis et al. 2011), and Random Forest (RF)(Beijma et al. 2014; Belgiu and Drăguţ 2016; Khatami et al. 2016).CART is an umbrella of Classification and regression task, and introduced by Breiman et al(Stutz and Taylor 2010). The goal of CART is to create a model that can predict the value of target variable by learning simple decision rules inferred from the data features. To obtain largest information gain, Gini index was used to select feature. Although the accuracy of this method is highest in these three conventional methods, CART is the only one white box model which means we can display graphically in a way that is easy to interpret.

The RF classifier is an ensemble classifier that uses a set of CARTs, and as shown in name, two random selection (feature and sample) was used in construction of trees. In the random selection of feature, out-of-Bag(OOB) data and permutation test can figure out the importance of each feature. About one thirds samples will not be used in the random selection of sample to train model, hence, this OOB data can be used to validate this model, which is different from other classifiers. Due to the outstanding characteristic that this model is computationally efficient, deal better with high-dimension data, and does not overfit, RF classifier has been successfully used in landcover mapping and other classification task(Belgiu and Drăguţ 2016).

SVM, introduced firstly by in 1995 by Cortes, is a high-performance classifier and aim to define a hyperplane to maximize the margin which is he distance between the separating hyperplane and closest sample (support vector). The trick of usage of non-linear kernel such as RBF (radial basis function) and Polynomial solve a non-linear classification problem. This method was particularly appealing in remote sensing imagery classification due to its ability of handling small training dataset and high accuracy. RF and SVM have been improved to be two of outperformance classification models and widely used in remote sensing classification.

All these three methods were implement in python with Scikit-Learn package(Pedregosa et al. 2012), detail parameters were shown in Table 3.

Table 3 Classification algorithms and parameter settings.

| Classifier | Parameters | Implementation | Remark |
|---|---|---|---|
| **RF** | Maximum tree: 300<br>Minimum leaf<br>sample size: 5 | scikit-learn | Data were scaled to [0, 255] before<br>training and classification |
| **SVM** | Kernel: RBF<br>C(cost): 300<br>gamma: 0.1 | | |

CART, SVM and RF

# Deep learning method

Three semantic segmentation deep learning methods were test in this work including Segnet, U-net and PSPnet. Their inputs, output and train parameters are listed in Table 1. The detailed architecture and feature introduction is as tableX. Meanwhile, we also got a merged result by averaging ensemble from the three individually trained models.

Table 1

| | SegNet | U-net | PspNet |
|---|---|---|---|
| **Input** | 256*256*7 | 256*256*7 | 512*512*7 |
| **output** | 256*256*7 | 256*256*7 | 512*512*7 |
| **Layer** | 94 | 103 | 293 |
| **Total params** | | | |
| | 29,461,736 | 34,615,752 | 51,818,112 |
| **Trainable params** | 29,445,848 | 34,601,736 | 51,759,872 |
| **Non-trainable params** | 15,888 | 14,016 | 58,240 |

# SegNet encoder-decoder network

SegNet is a combination of a fully convolutional network and an encoder-decoder architecture. The input image is first passed through a sequence of convolutions, ReLU, and max-pooling layers. During max-pooling, the network tracks the spatial location of the winning maximum value at every output pixel. The output of this encoding stage is a representation with reduced spatial resolution. That "bottleneck" forms the input to the decoding stage, which has the same layers as the encoder but in reverse order. Max-pooling layers are replaced by un-pooling, where the values are restored to their original locations, and the convolution layers are then used to interpolate the higher-resolution image. Because the network does not have any fully connected layers (which consume >90% of the parameters in a typical image-processing CNN), it is much leaner. SegNet is therefore highly memory efficient and comparatively easy to train(Kendall et al. 2015).

The SegNet architecture, which fully considers training time and memory allocation factors, has been shown to be efficient for image semantic segmentation and is significantly smaller and faster than competing architectures. The SegNet model was designed for RGB-like image segmentation. Meanwhile, SegNet inputs can be multi-channel and are particularly suitable for unbalanced labels(Badrinarayanan et al. 2015). Therefore, the model is particularly suitable for remote sensing image classification. The Development Seed company created the Skynet project. Skynet can quickly analyze massive amounts of satellite imagery using the SegNet architecture deep learning method and open data (OpenStreetMap)(Seed 2017).

SegNet has several attractive properties: (1) it only requires the forward evaluation of a fully learned function to obtain smooth label predictions; (2) with increasing depth, a larger context is considered for pixel labeling, which improves accuracy; (3) the effects of feature activation(s) in the pixel label space at any depth are easily visualized; and (4) SegNet inputs can be any arbitrary multichannel image or feature map suitable for remote sensing data.

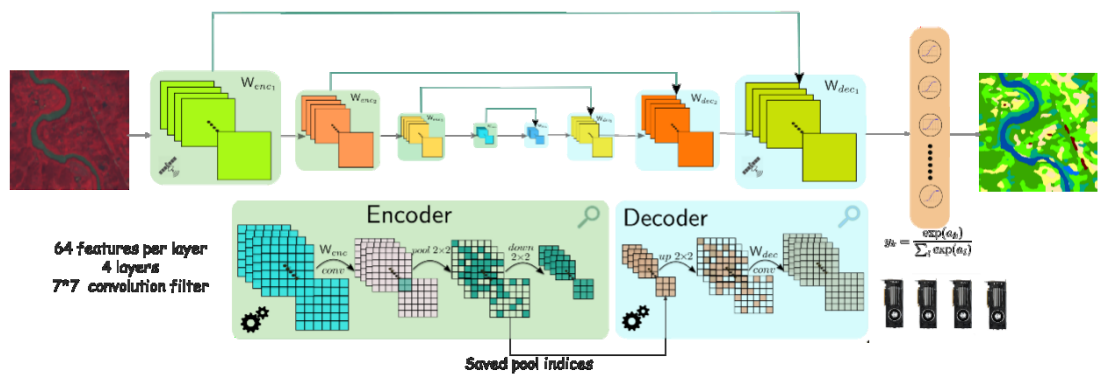

Figure 7 An illustration of the SegNet architecture. The 4-layer SegNet takes in multi-band imagery and performs feed-forward computations to obtain pixel-wise labels. A decoder upsamples its input using the transferred pool indices from its encoder to produce (a) sparse feature map(s)(Badrinarayanan et al. 2015).

## U-Net

The U-Net was initially published for bio-medical segmentation, the utility of the network and its capacity to learn from very little data, it has found use in several other fields satellite image segmentation and also has been part of winning solutions of many kaggle contests on medical image segmentation. The name was like it's structure 'U'. U-Net simply concatenates the encoder feature maps to up sampled feature maps from the decoder at every stage to form a ladder like structure. Meanwhile, the architecture by its skip concatenation connections allows the decoder at each stage to learn back relevant features that are lost when pooled in the encoder. The architecture was simple and efficiency which consists of a contracting path to capture context and a symmetric expanding path that enables precise localization.

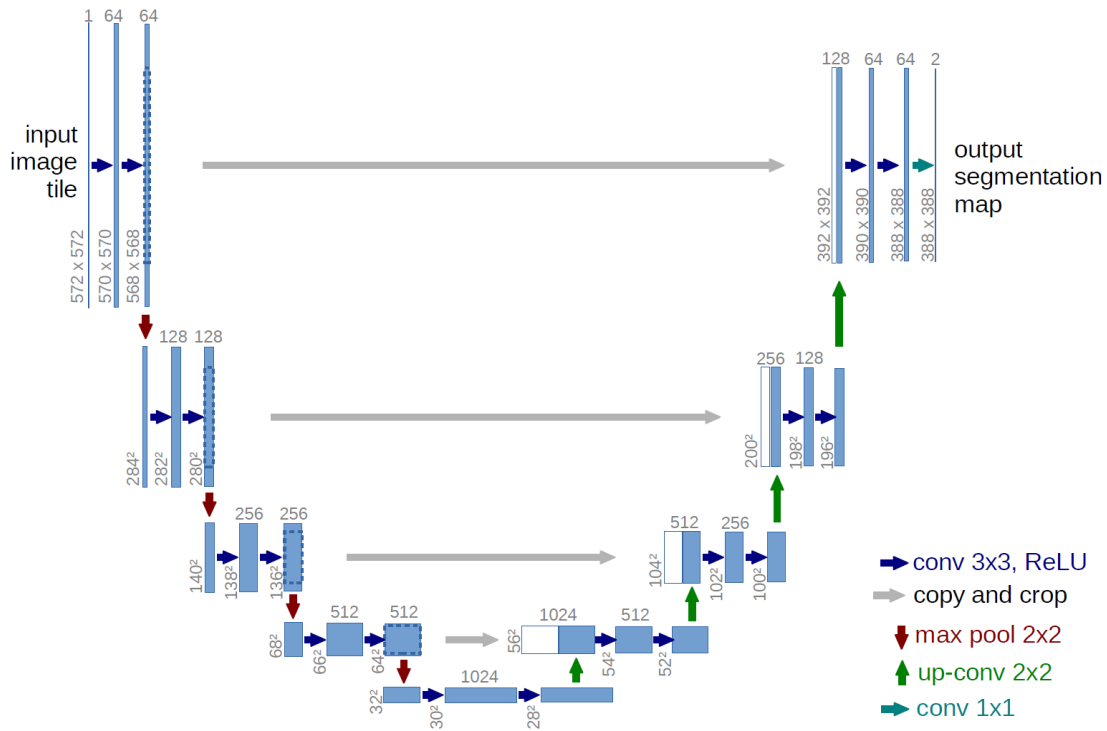

Figure 8

# PspNet

PSPNet modifies the base ResNet architecture by incorporating dilated convolutions and the features, after the initial pooling, is processed at the same resolution (1/4th of the original image input) throughout the encoder network until it reaches the spatial pooling module. Meanwhile, it added the auxiliary loss at intermediate layers of the ResNet to optimize learning overall learning. Spatial Pyramid Pooling at the top of the modified ResNet encoder to aggregate global context.

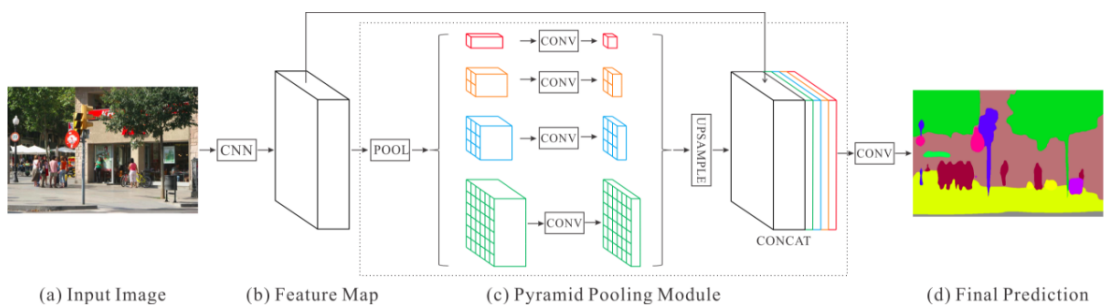

(a) Input Image        (b) Feature Map        (c) Pyramid Pooling Module        (d) Final Prediction

## Model training and predict

 For training the deep learning model, the Keras based on Tensorflow was used to build and training the model. For Segnet and U-net model, the Adam was selected as the optimizers and the learning rate was set as 1e-5(Kingma and Ba 2014). For PSPnet, the stochastic gradient descent (SGD) was set as the optimizer, the learning rate was set as 0.05 and the momentum was set as 0.9. the categorical cross entropy was set as the loss function in the training.

The image was divided into 256*256 or 512*512 pieces due to the limited of deep learning method. After the predict, we combined all the pieces together. To avoid possible seams between pieces edge. An expansion prediction was used in this work. As shown as the Figure 9, only the central area for each piece was used as the result.

In this work, 60% of data was used to training the model with 20% were validate and adjust the model training rate. The left 20% were used for validation.

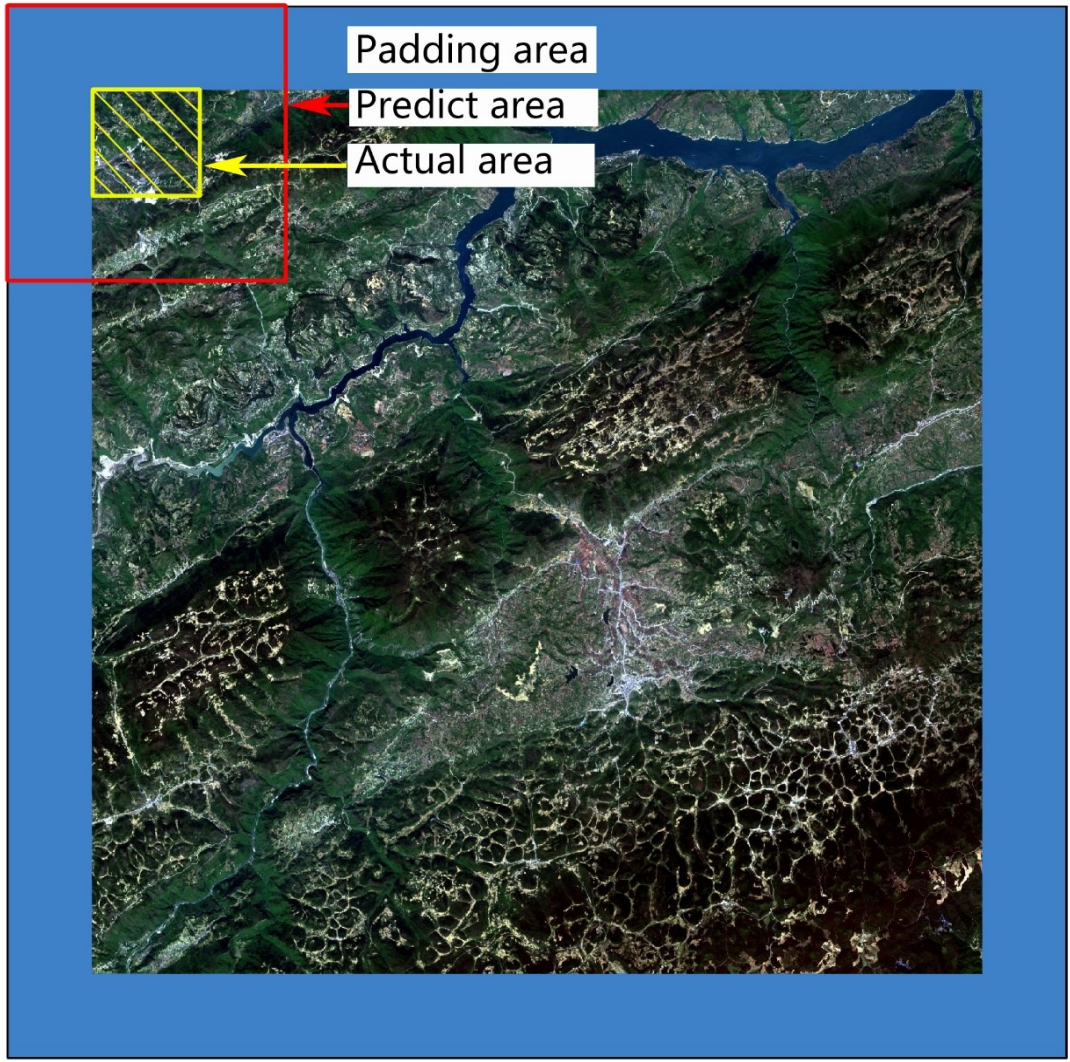

Figure 9

## Aaccuracy assessment (recall ,precision and F1)

The performance of the different approaches was assessed by two complementary criteria, namely the accuracy assessment and across-site robustness. Two different metrics, derived from the confusion matrix, were selected for the overall accuracy (OA) assessment. The OA evaluated the overall effectiveness of the algorithm, while the F1-score measured the accuracy of a class using the precision and recall measures. The study establishes error matrices in each province that provides user's (UA), producer's (PA), and overall accuracies (OA) as following equations:

$$OA = \frac{S_d}{n} \times 100\%$$

$$UA = \frac{X_{ij}}{X_j} \times 100\%$$

$$PA = \frac{X_{ij}}{X_i} \times 100\%$$

$$F1_{score} = \frac{UA \times PA}{UA + PA} \times 2$$

where $S_d$ is the total number of correctly-classified pixels, n = total number of validation pixels, $X_{ij}$ = observation in row i column j in confusion matrix; $X_i$= marginal total of row I and $X_j$ = marginal total of column j in confusion matrix.

# Results and Discussion

## 1.The importance of dem and slope

The accuracy of three convolutional pixel-based methods was shown in Table 2. For shrub, grassland and cropland. The SVM and RN classifier are difficult to identified them. On the one hand, the spectral information was highly similar among different vegetation classes (Figure 13). It was difficult to distinguish different vegetation types, especially forests, shrub, grassland and cropland without resorting to multi-temporal imagery.

We compared two combinations of features, one is only the five bands in rapideye imagery that is blue band, green band, red band, red edge band and near inferred band, the other one contains DEM and slope besides the five bands of rapideye. With added DEM and slope, the CART and RF method overall accuracy increased 0.03 and the F1 score of every class increased significantly. At the meanwhile, due to the characteristic that SVM was decided by several supporters, and was not sensitivity to feature dimension( ), the accuracy of SVM didn't increase. In these three methods and two feature combinations, RF with 7band feature has the highest accuracy. Compared with the other two methods, RF often outperformed, which was coincide with previous research (引用).

The comparison of 5band feature and 7band feature using RF method was shown in Figure 10, the F1 Score of each class has increased significantly. Using RF method with 7band rather than 5band, F1 score of shrubland and cropland increased 0.4, which is the highest increase, Tree Cover and grass land increased 0.3, another two landcover classes (Artificial surface and water body) increased 0.2. The 5 bands , DEM and Slope range of each class was shown in Figure 11 and Figure 12, respectively. From that, it can be concluded that the different landcover incline to different elevation: The elevation of Tree cover was highest followed by shrubland, grassland, farmland, artificial surface and water land. Water land most distributed in flat area and tree cover most located on the gradient like mountains. So, DEM and Slope was important to classification task, which was confirmed by the result that DEM was the most important feature in RF feature importance figure (Figure 11).

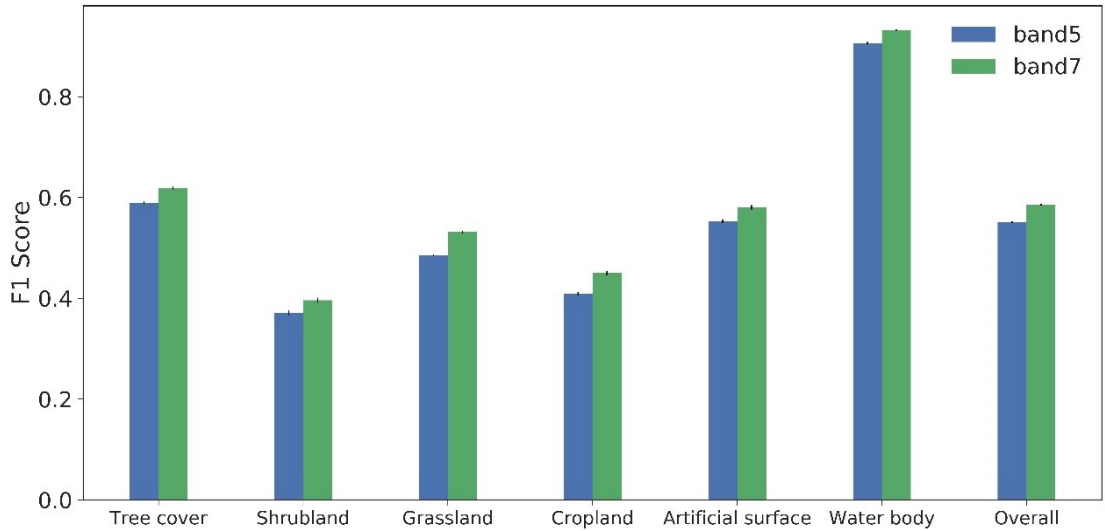

Figure 10 F1 Score of each class used RF clasifier with 5band feature and 7band feature

Table 2 F1 score of three pixel-method and two feature combinations.

| *Category* | CART | | SVM | | Random Forest | |
|---|---|---|---|---|---|---|
| | 5-band | 7-band | 5-band | 7-band | 5-band | 7-band |
| Tree cover | 0.47 | 0.5 | 0.57 | 0.55 | 0.59 | **0.62** |
| Shrubland | 0.31 | 0.32 | 0.37 | 0.37 | 0.37 | **0.4** |
| Grassland | 0.36 | 0.4 | 0.5 | 0.49 | 0.49 | **0.53** |
| Cropland | 0.32 | 0.34 | 0.42 | 0.41 | 0.41 | **0.45** |
| Artificial surface | 0.44 | 0.45 | 0.54 | 0.51 | 0.55 | **0.57** |
| Water body | 0.86 | 0.9 | 0.9 | 0.91 | 0.91 | **0.93** |
| Overall | 0.46 | 0.49 | 0.55 | 0.54 | 0.55 | **0.58** |

*5-band means blue band, green band, red band, red edge band and near inferred band, 7-band contains 5-band, DEM and slope.

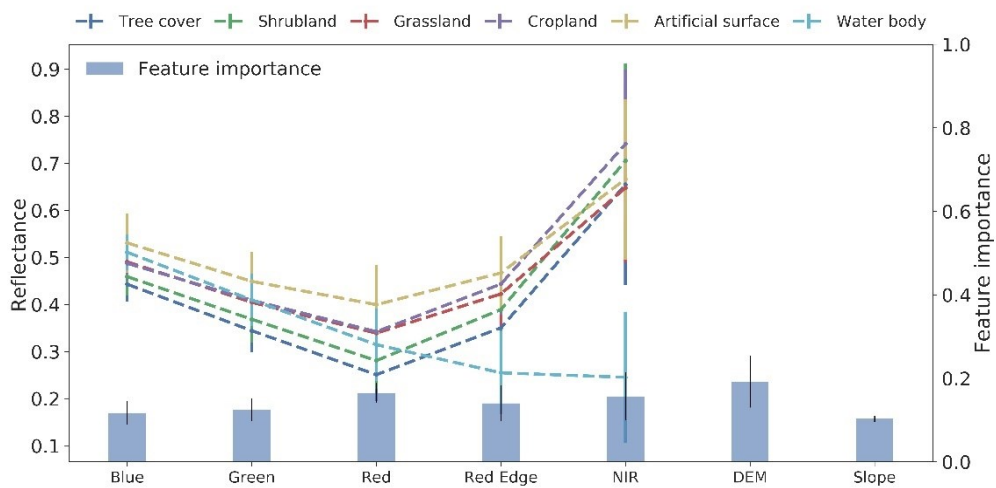

Figure 11 The 5 bands range of each class and the feature importance in RF method.

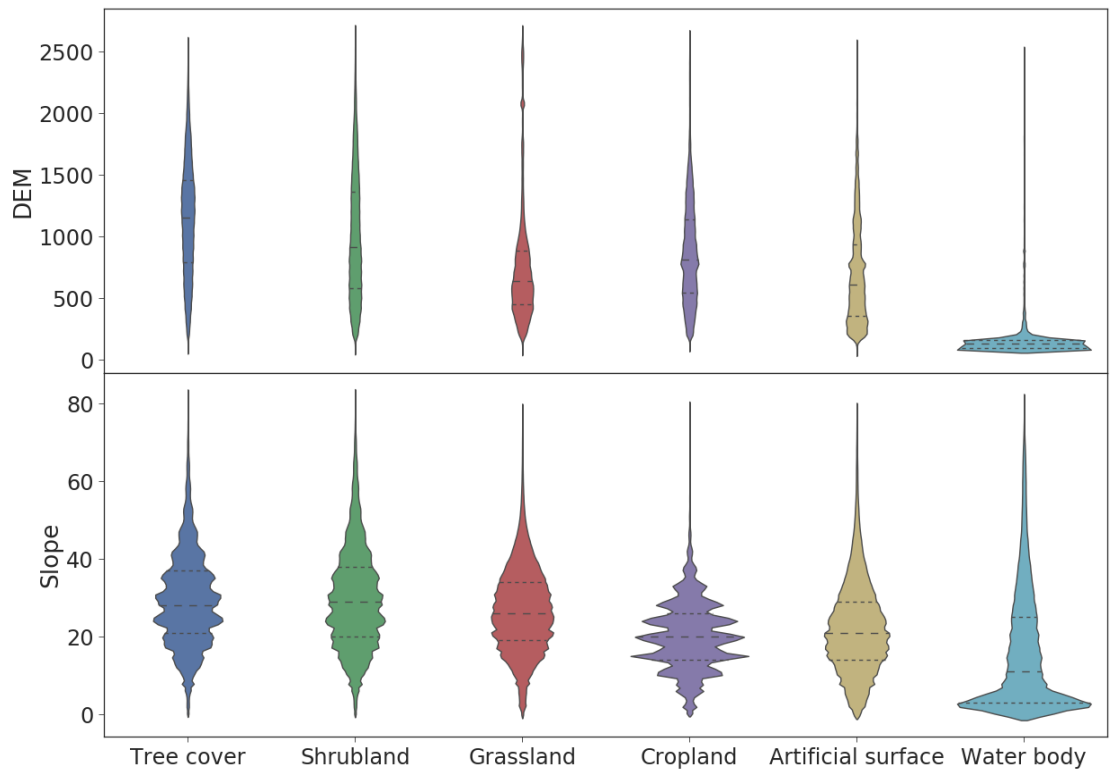

Figure 12 The DEM and slope range of each class.

## 2. The classification results across deep learning methods

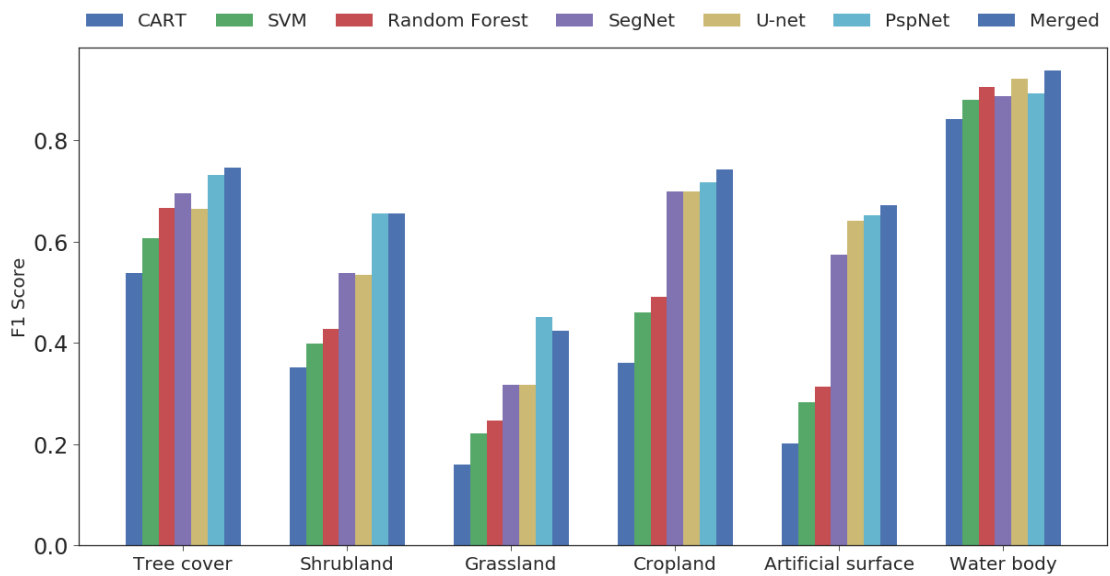

Figure 13 the F1 score for each method across categories

Pervious result demonstrates the topographic information (DEM and slope in this paper) can improve the accuracy for each method. We test three deep learning methods including SegNet, U-net, PspNet in this section. The F1 score for each method across categories are shown in Figure 13. For pixel-based method, the random forest has the best performance for all categories. Meanwhile, in most instances, the results based on deep learning methods have better performance. The SegNet architecture was originally designed for classifying road scenes and was first applied to the classification of high-resolution remote sensing images, due to its efficient and concise architecture, achieving good results. Nicolas Audebert used the SegNet architecture with a fusion table to classify earth observation data (ISPRS 2D Semantic Labeling Challenge Dataset at 0.125m resolution) into 5 types, achieving an accuracy of 89.8%(Audebert et al. 2016). The UNET which has a similar structure to SegNet, has also been widely applied in Kaggle competition for remote sensing application and has been proved better than U-net. Kohei Ozaki and Артур Кузин choose UNET as their winning Solution for the Spacenet Challenge and Defence Science and Technology Laboratory (Dstl) Satellite Imagery Feature Detection (Kaggle)(SpaceNet 2017; Kaggle 2017). In this work, we firstly test these two popular architectures in landcover mapping based on 5m resolution remote sensing images. The F1 score accuracy based on Segnet and U-net's results are better than pixel-based method, especially on Shrubland, Grassland, Cropland and Artificial surface. For most categories, the two methods have similarly performance and the U-net is slightly better with SegNet. The PspNet, which achieves state-of-the-art performance on various datasets including CityScapes, ADE20K and Pascal VOC 2012, has the largest number of layers in our work. The F1 score accuracy proves the superiority of this model. especially in shrubland and grassland, the accuracy based on PspNet was higher than other models significantly. Model ensemble was a common trick that can improve accuracy efficiency. We average ensemble the Segnet, U-net and PspNet results in this work, the merged result shows the best performance in most categories except grassland. Figure 14 shows the classification confusion matrix for Segnet, U-net, PspNet and the merged result. The water body is the most easily distinguished category due to its spectral properties and topographic information (Figure 11). The tree cover, shrubland, grassland, cropland and artificial surface has similar spectral properties and topographic information which made it difficult to distinguish. The result on grassland has the poorest performance. The vast majority of pixels in grassland are assigned to shrubland and cropland. Meanwhile, there are many pixels in artificial surface was assigned to cropland. The reason is the Wushan is a farming-based county where most cropland was distributed around residential areas (Figure 2) which made it easily confused. Another reason is most images is acquired in summer and autumn when most cropland is near harvest or after harvest. Some bare cropland after harvest was assigned as artificial surface.

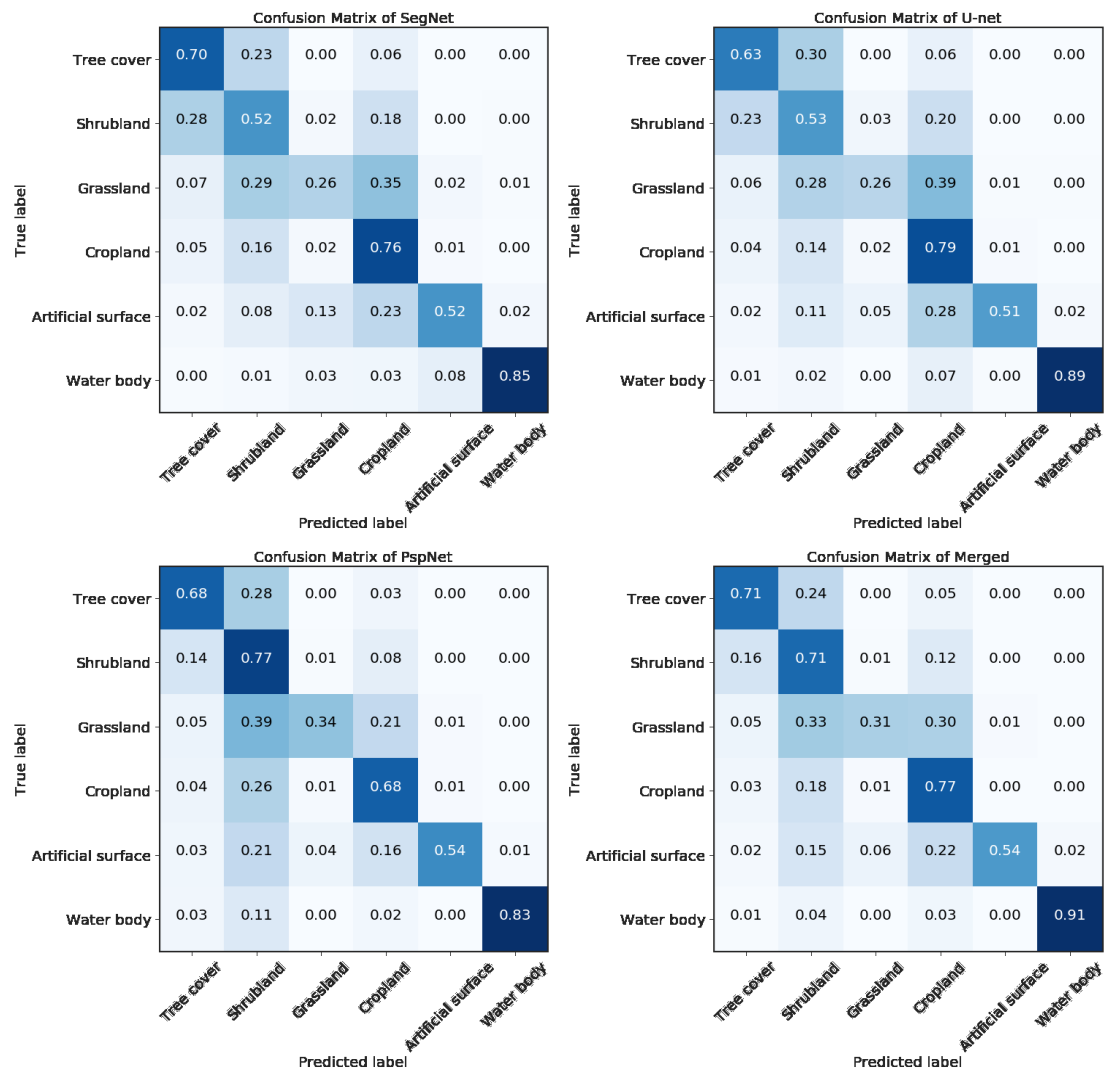

Figure 14 Confusion matrix of the classification results achieved by the Segnet, U-net, PspNet and merged methods based on the validate datasat in 2012.

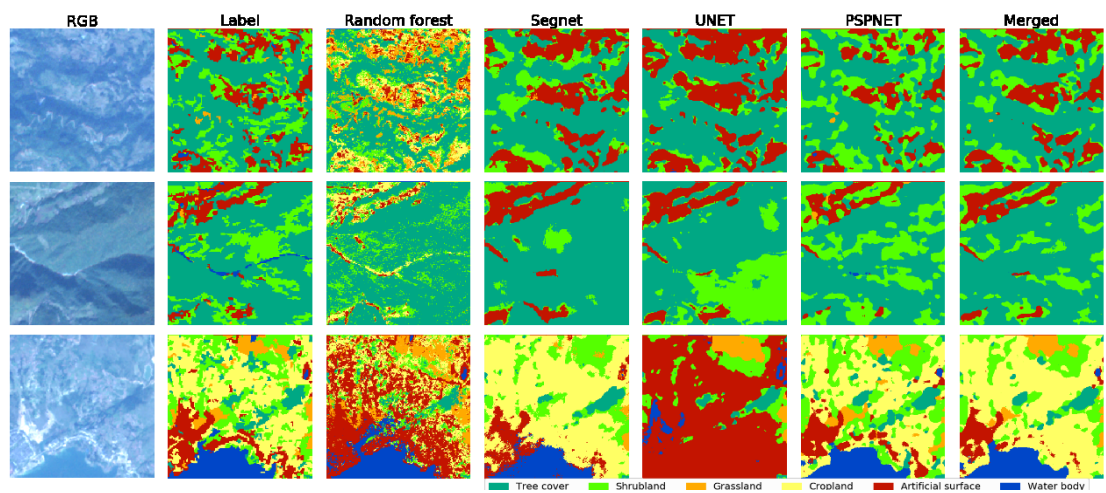

Figure 15 the results comparison of the Random forest for pixel based methods, Segnet, UNET, PSPNET, merged for deep learning method and the label made by object-based approach.

The object-based approach to remote sensing analysis and target identification has received considerable attention accompanied by the release of medium to high resolution data. This approach works similar to the human eye recognition that colors, geometric features, spatial relations, and scale effects are comprehensively considered in target recognition. Another advantage of the object-based approach is its ability to integrate complex spatial information and to reduce the 'salt-and-pepper effect'(Zhang et al. 2014). The label in 2012 was based on an object-based approach and operated in eCognition software. This method divides the landcover work into two steps: segmentation and classification. The segmentation takes a lot of computing resources and time consumption and the classification still requires human involvement including choosing classifier and adjust the parameters. Figure 15 shows the results comparison of the Random forest for pixel based methods, Segnet, UNET, PSPNET, merged for deep learning method and the label made by object-based approach. As previous research has confirmed, the result based on pixel-based method (random forest result in this figure) appears messy visually in medium to high resolution(Xiong et al. 2017). Our results also proves that the pixel-based method preformace poor at 5m resolution compared with object-based approach visually. However, the deep learning methods which combining segmentation and classification shows visually similar result as the object-based method. Once the training is completed, the deep learning method does not require manual operation, which can greatly save time consumption. The training time for deep learning are still long compared with other methods which led to the generalization and adaptability of the model is crucial. we will evaluate the generalization and robustness of deep learning methods in the next section.

## 3.Robustness to uncertainties data in 2016

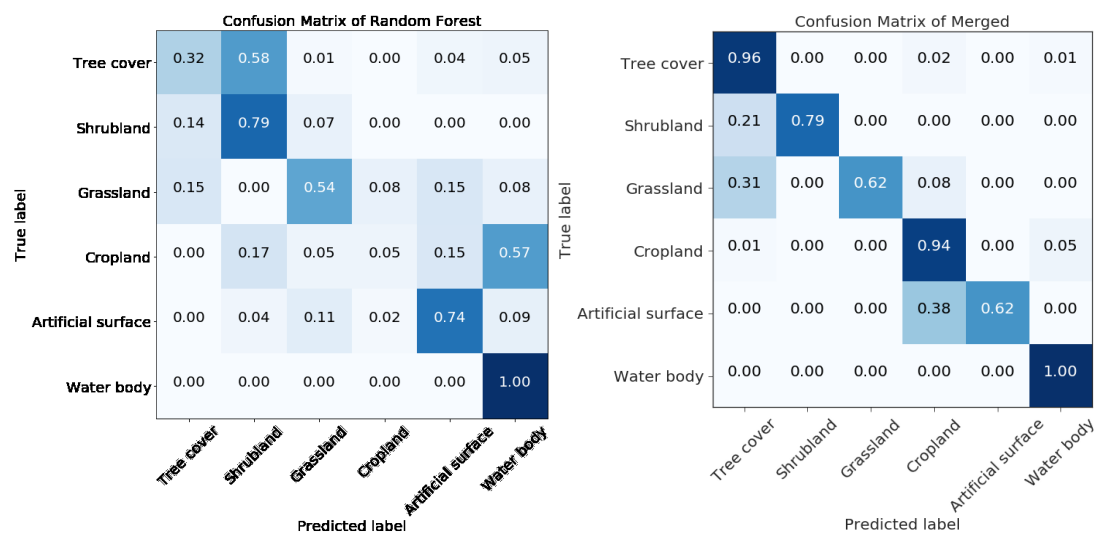

Figure 16 Confusion matrix of the classification results achieved by the Randon forest, deep learning methods merged result based on the 2016 datasat in Wushan county.

Table 3

| Category | Count | precision | recall | f1-score |
|----------|-------|-----------|--------|----------|

|  |  | RF | Merged | RF | Merged | RF | Merged |
| --- | --- | --- | --- | --- | --- | --- | --- |
| Tree cover | 81 | 0.87 | 0.91 | 0.32 | 0.93 | 0.47 | 0.96 |
| Shrubland | 14 | 0.15 | 1 | 0.79 | 0.88 | 0.25 | 0.79 |
| Grassland | 13 | 0.39 | 1 | 0.54 | 0.76 | 0.45 | 0.62 |
| Cropland | 80 | 0.67 | 0.78 | 0.05 | 0.85 | 0.09 | 0.94 |
| Artificial surface | 47 | 0.67 | 1 | 0.74 | 0.76 | 0.71 | 0.62 |
| Water body | 1 | 0.02 | 0.17 | 1 | 0.29 | 0.04 | 1 |
| Overall | 236 | 0.69 | 0.89 | 0.36 | 0.86 | 0.37 | 0.86 |

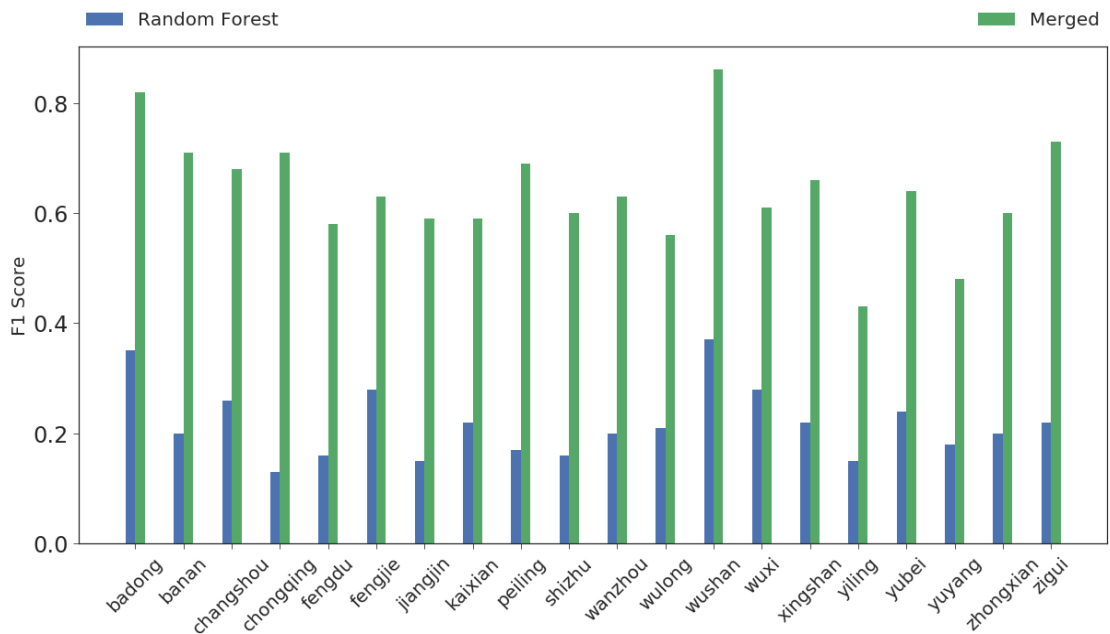

Figure 17

In this section, 4,174 fields ground truth points obtained in 2016 were used to evaluate the generalization and adaptability of the classifier model. the random forest which performance best in pixel-based methods and deep learning merged model were used and the models were trained by 2012 data in the last section. The confusion matrixs and the precision, recall and F1 score for each category of the two classifier results in Wushan county were shown in Figure 16 and Table 3. The result based on deep learning methods merged was significantly better than random forest classifier. Meanwhile, we also test these two models on other 19 county in TGRA. The merged classifier based on deep learning methods was also better than the random forest classifier (Figure 17). The F1 scores of merged method in all 20 countys were between 0.48 to 0.86 which were close to the accuracy based on validate data in 2012. However, the performance of random forest on 2016 data were significantly lower than the result based on 2012.

Of course, the pixel-based method like CART, SVM and random forest were easy to training on smaller dataset with point label. We can retraining the model on limited 2016 datasat for better performance. However, deep learning semantic segmentation models are often improved by the availability of massive datasets and by expanding model depth and parameterizations(Lecun et al. 2015). The labeling data limitations are a crucial factor for the development. In compute visualization area, IMAGENET and COCO dataset provide enough materials and still increased. For remote sensing classification in high and very high resolution, Audebert and Jason Remillard tried get labeling dataset from Volunteered Geographic Information (VGI) including OpenStreetMap(Audebert et al. 2017). The DigitalGlobe also publish a corpus of commercial satellite imagery and labeled training data to use for machine learning research(AWS 2017). Which has greatly improved the development of deep learning semantic segmentation model.

## Conclusion

In this paper, we first tested the deep learning semantic segmentation classifier on land cover mapping in the TGRA area with 7 categories at 5m resolution. The mission was very challenging. Currently, there are no automatic classification methods that can effectively complete this task. In the past the object-oriented method and visual interpretation were used to obtain high accuracy, but it requires large amounts of manpower. In this study,

1. DEM and slope could add accuracy for pixed-based method, also works on DL method.
2. DL method has a better segment result with medium to high resolution.
3. DL method has a better accuracy and robust for predict data.

## Acknowledgements

This paper was supported by the National Key R&D Program of China (2016YFA0600301) and National Natural Science Foundation for Young Scholars of China (Grant No. 41501474), Thanks go to the anonymous reviewers for reviewing the manuscript and providing comments to improve the manuscript.

## Author Contributions:

Xin Zhang contributed to the research experiments, analyzed the data, and wrote the paper. Bingfang Wu conceived the experiments, and responsible for the research analysis. Zhu Liang collected and pre-processed the original data. All the co-authors helped to revise the manuscript.

## Conflicts of Interest:

The authors declare no conflict of interest.